\documentclass[iicol,sn-basic]{sn-jnl}

\usepackage{caption}
\usepackage{booktabs}
\usepackage[para,online,flushleft]{threeparttable}

\usepackage[marginal]{footmisc}
\renewcommand{\thefootnote}{}

\jyear{2023}

\theoremstyle{thmstyleone}

\theoremstyle{thmstyletwo}

\theoremstyle{thmstylethree}

\newcommand{\etal}{\textit{et al}. }

\begin{document}

\title[ADR]{Adaptive Discriminative Regularization for Visual Classification }

\author[1]{\fnm{Qingsong} \sur{Zhao}}\email{qingsongzhao@tongji.edu.cn}
\equalcont{These authors contributed equally to this work.}

\author[2]{\fnm{Yi} \sur{Wang}}\email{wangyi@pjlab.org.cn}
\equalcont{These authors contributed equally to this work.}

\author[1]{\fnm{Shuguang} \sur{Dou}}\email{2010504@tongji.edu.cn}
\author[3]{\fnm{Chen} \sur{Gong}}\email{chen.gong@njust.edu.cn}
\author[1]{\fnm{Yin} \sur{Wang}}\email{ yinw@tongji.edu.cn}
\author*[1]{\fnm{Cairong} \sur{Zhao}}\email{zhaocairong@tongji.edu.cn}

\affil[1]{\orgdiv{Department of Computer Science and Technology}, \orgname{Tongji University}, \orgaddress{\city{Shanghai}, \postcode{201804}, \country{China}}}

\affil[2]{\orgname{Shanghai AI Laboratory}, \orgaddress{\city{Shanghai}, \postcode{200232}, \country{China}}}

\affil[3]{\orgdiv{Key Laboratory of Intelligent Perception and Systems for High-Dimensional Information, School of Computer Science and Engineering, Ministry of Education}, \orgname{Nanjing University of Science and Technology}, \orgaddress{\city{Nanjing}, \postcode{210094}, \country{China}}}

\abstract{
How to improve discriminative feature learning is central in classification. Existing works address this problem by explicitly increasing inter-class separability and intra-class compactness, whether by constructing positive and negative pairs for contrastive learning or posing tighter class separating margins. These methods do not exploit the similarity between different classes as they adhere to independent identical distributions assumption in data. In this paper, we embrace the real-world data distribution setting that some classes share semantic overlaps due to their similar appearances or concepts. Regarding this hypothesis, we propose a novel regularization to improve discriminative learning. We first calibrate the estimated highest likelihood of one sample based on its semantically neighboring classes, then encourage the overall likelihood predictions to be deterministic by imposing an adaptive exponential penalty. As the gradient of the proposed method is roughly proportional to the uncertainty of the predicted likelihoods, we name it adaptive discriminative regularization (ADR), trained along with a standard cross entropy loss in classification. Extensive experiments demonstrate that it can yield consistent and non-trivial performance improvements in a variety of visual classification tasks (over 10 benchmarks). Furthermore, we find it is robust to long-tailed and noisy label data distribution. Its flexible design enables its compatibility with mainstream classification architectures and losses.
}


\maketitle
\footnotetext{
This work is being submitted to a journal.
}

\begin{figure*}[!tb]
\begin{center}
  \includegraphics[width=1.0\linewidth]{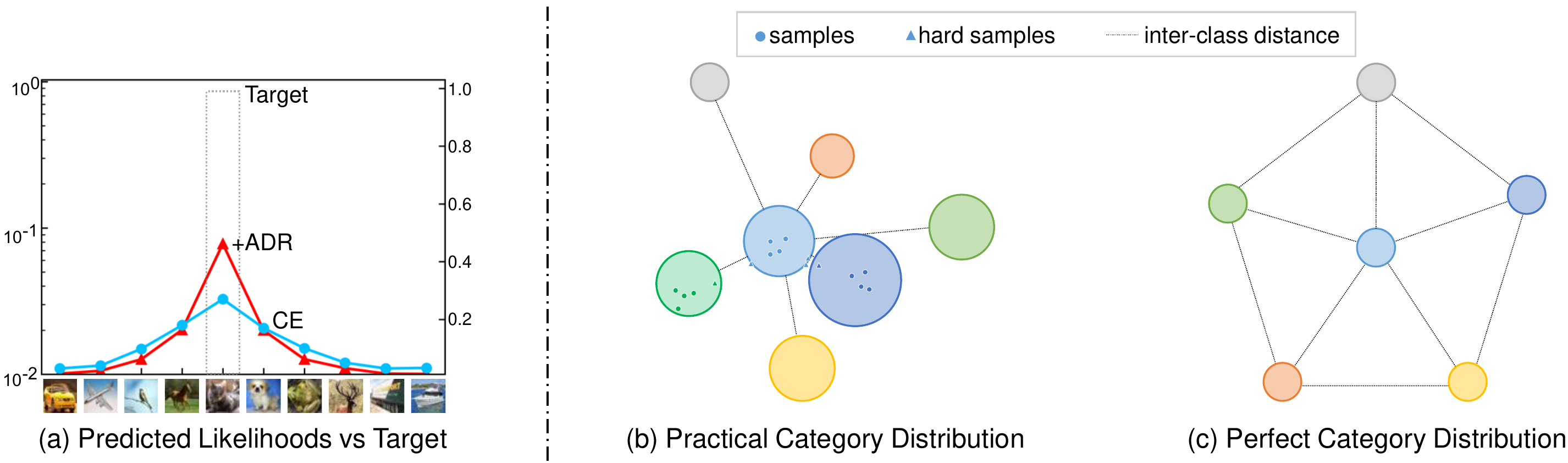}
  \caption{Different predicted logit distributions are plotted in (a).
  Different data distributions are drawn in (b) and (c). 
  When collecting data in a wild, one often expects the distance between clusters to be equal i.e., (c) an ideal category distribution.
  But the most cases, we collected the dataset can be represented as (b), in which the clusters vary in variance and inter-class distance (various sizes of circles indicate different variances). 
  The hard samples at the edge of the clusters contribute significantly to the decision surface (see \cite{Forgetting}), while they are more likely to make ambiguous predictions as the blue curve drawn in (a). 
  }
  \label{category_distribution}
\end{center}
\end{figure*}

\section{Introduction}\label{sec:Introduction}
Visual classification is one of the fundamental research topics in machine learning and computer vision. Typically, it transforms high-dimensional visual signals (e.g., images, videos, etc.) into the corresponding latent features, and then differentiates them into different classes. With the advance of deep learning and the availability of big data, visual classification makes significant leaps in its both theories and practices, widely employed in face recognition, object detection, and so on.

In visual classification studies, to improve its discrimination ability, efforts have been made to increase inter-class separability and intra-class compactness. 
Existing methods (\cite{hadsell2006dimensionality, liu2016large, zhu2019large,  sun2020circle}) are either based on positive and negative sample pairs or large-margin softmax, the former favors representational learning and the latter is an optimization of the decision surfaces. 
Specifically, the L-softmax (\cite{liu2016large}) and its variants (\cite{liu2017sphereface, wang2018cosface, deng2019arcface}) reinforce the deep neural network learning a bigger margin around the separating hyperplanes by factorizing the cosine similarity into amplitude and angular.
This idea is equivalent to making hyperplanes with a larger margin than the original ones driven by a vanilla softmax (aka cross entropy) loss.
Both these two types of methods usually work decently on fine-grained (e.g., LFW (\cite{huang2008labeled})) or small-scale visual classification dataset (e.g., CIFAR-10 (\cite{krizhevsky2009learning})), 
while bringing trivial benefits to large-scale discrimination tasks.
We suppose these lifting optimization low-bound methods could accelerate large-scale classification training but hardly improve its performance.
Because real-world data are not distributed ideally, hard examples (usually found in large-scale datasets) hinder the effective training of large class margins.

Specifically, training and testing data in the existing visual classification are supposed to be in independent identical distributions (i.i.d.). In this scenario, training a classification model by minimizing the cross entropy between the predicted likelihood and the given ground truth can lead to a discriminative representation, guaranteed by maximum likelihood estimation.
This usually does not hold true, as several defined categories share similar concepts (e.g., cats and tigers have similar visual appearance) and some ones have large intra-class dispersion (i.e. one single cluster has different similarity to the others as given in Fig. \ref{category_distribution} (b)).
We suppose that can be a big challenge for the traditional maximum likelihood estimation training strategy built upon i.i.d, see \cite{arora2018stronger} and \cite{banburski2021distribution} for similar statements.
For this purpose, we propose a new classification regularization on the estimated likelihoods by exploiting inevitable data dependence. 

\begin{figure*}[!tb]
\begin{center}
\includegraphics[width=\textwidth]{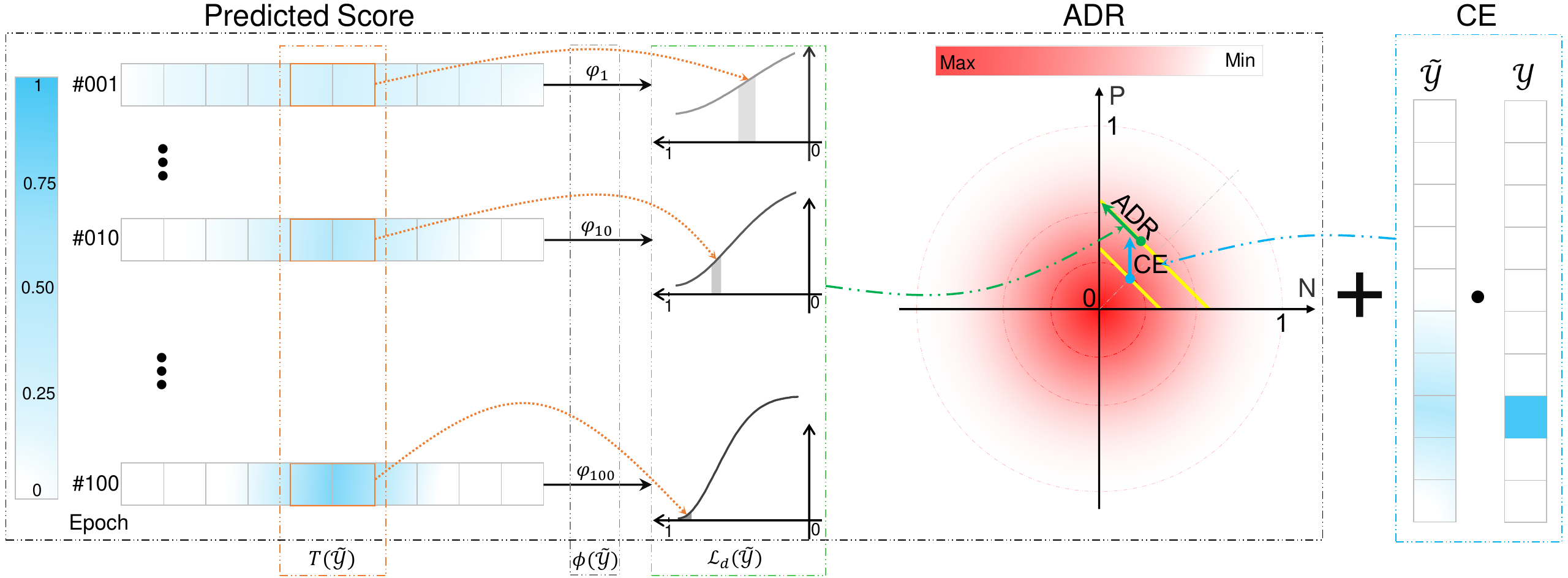}
\end{center}
  \caption{The conceptual illustration of how training a supervised model with our ADR in a toy experiment (details see Sec. \ref{Discussion}). 
  Minibatch images are fed into the backbone and softmax to obtain the predicted logits ${\tilde {\cal Y}}$. 
  At different training phases, we capture the predicted logits, normalize them according to their confidence, and plot the change curve of ADR (the gray shaded area indicates the possible range of Top-1 logits).
  The cross-entropy loss pushes the optimization direction (the blue arrow) to the positive side at the beginning, while our ADR enlarges the separability between classes and makes the optimization fall to one side further (the green arrow). }
  \vspace{-0.05in}
\label{fig:pipline}
\end{figure*}

Due to the pervasive inter-class similarities and intra-class dissimilarity, as shown in Fig. \ref{category_distribution} (a) the predicted likelihoods (i.e. predicted logits from softmax) tend to show a smooth distribution instead of a spiky one, contradicting the initial data assumption that each example has only one label.
Thus, we propose an adaptive discriminative regularization to encourage the predicted likelihoods to be deterministic, and stabilize such optimization procedure by controlling gradient magnitude according to the certainty of likelihoods.
Specifically, as shown in Fig. \ref{fig:pipline},
we firstly calibrate the predicted maximum likelihoods of one sample by its semantically similar classes,
then we exert a discriminative constraint on the predicted likelihoods based on a normalized exponential function,
that does not only makes the corresponding gradients adaptive to the confidence of predicted likelihoods (i.e. high deterministic likelihoods give a small gradient magnitude, while low deterministic likelihoods give a big one), 
but also is optimization-friendly.

\begin{figure*}[tb]
	\includegraphics[width=\textwidth]{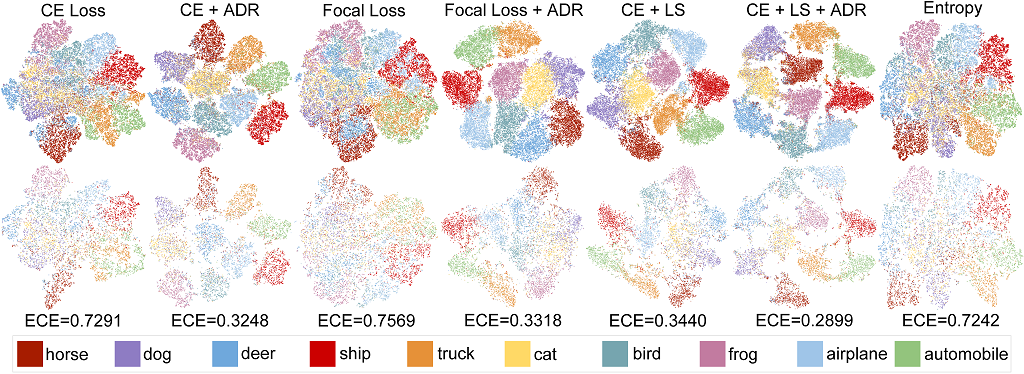}
	\caption{Visualization of classifier layer's features. The first row comes from a training set, the second row is from a validation set. To be convincing, we plotted the entire sample. The first five columns represent the features extracted from the model which is trained by CE loss, CE w/ ADR, CE w/ LS, CE w/ LS w/ ADR, and CE w/ Entropy loss respectively. The Expected Calibration Error (ECE) on the CIFAR-10 validation set is reported (below) too. Lower is better. }
	\label{fig:tsne}
\end{figure*}

We validate the effectiveness of our hypothesis and the corresponding regularization by visualizing the features learned by different optimization targets, as shown in Fig. \ref{fig:tsne}. We quantify how well the trained classification models perform by \cite{guo2017calibration} the expected calibration error (ECE). Compared to other methods, the proposed ADR can achieve a lower ECE value easily.
Note that the inter-class distance with our ADR is larger than that with the existing baselines.
Intuitively, it can bridge the gap between the estimated likelihood and the practical data distribution (i.e. it may have a better separating hyperplane).

We thoroughly discuss the properties of our proposed ADR theoretically.
More importantly, we empirically validate its effectiveness in multiple mainstream classification benchmarks with various settings in Sec. \ref{Experiments}. 
Extensive experiments show our ADR improves current mainstream model architecture with standard classification optimizations non-trivially. 
Also, its universal properties make the trained model robust to long-tailed and noise-labeled validation data.
In summary, our contributions are summarized as follows: 
\begin{itemize}
	\item We design a new discriminative regularization approach for the supervised visual classification to enlarge the inter-class distance. It is relatively orthogonal to various discriminative optimization targets as it can further improve existing baselines non-trivially. 
	\item We demonstrate that the proposed ADR is compatible with backbones in convolutional neural network (CNN), transformer, and multilayer perceptron (MLP) architectures, and it exhibits robustness against noisy labels and long-tailed distributions.
\end{itemize}

\section{Related Work and Preliminaries}\label{sec2}
Existing works such as contrastive loss (\cite{hadsell2006dimensionality}), triplet loss (\cite{schroff2015facenet}), L-Softmax loss (\cite{liu2016large}), SphereFace (\cite{liu2017sphereface}), Cosface (\cite{wang2018cosface}), Arcface (\cite{deng2019arcface}), online label smoothing (OLS \cite{zhang2021delving}) and circle loss (\cite{sun2020circle}) have been proposed to enhance the performance of traditional softmax cross entropy loss.
In this section, we will briefly overview these methods respectively according to their motivations.
Specifically, this work is inspired by superset label learning, we will introduce the definition and some research works on it at the end.

\paragraph{Model Regularization. } 
In deep learning, many strategies are known collectively as \textit{regularizations}.
In order to reduce the validation error, those strategies often trade off the increase in training error.
For example, to alleviate the issue of over-fitting, label smoothing (\cite{szegedy2016rethinking}) avoids the search for exact likelihoods. 
It computes cross entropy with a weighted mixture of uniform distribution of targets (i.e. injecting noise into the targets). 
label smoothing (LS) chooses to discard particularly difficult samples in exchange for gathering simple samples well.
In brief, label smoothing should be more friendly to noise samples compared with focal loss (\cite{lin2017focal}). 
In \cite{muller2019does},  Muller \etal discussed why and when label smoothing should work, and demonstrated that label smoothing implicitly calibrates learned models.
Arguably, the accuracy improvement is not obvious or even decreases if using relatively small networks and face verification tasks with LS.
But in the same case, our method still works (please turn to Sec. \ref{sec:Image Classification} for details).

\paragraph{Discriminatory Feature Learning. }
For the same motivation (i.e. learning discriminatory features), 
existing methods are either designed to increase the learning difficulty of separating hyperplanes or require positive and negative samples as training.
For example, contrastive loss requires the same class features to be as similar as possible, yet the distance between different class features is larger than a margin.
And the triplet loss requires $3$ input samples at a time and maximizes the distance between the anchor and a negative sample.
But they all require a carefully designed pair selection procedure.
By contrast, the L-Softmax loss was firstly proposed in a novel view of the cosine similarity to learn discriminative features, and bring a series of extension researches (\cite{wang2018cosface,liu2017sphereface,deng2019arcface}).
For example, despite the similarity between ArcFace and previous works, it has a better geometric attribute.
However, all those methods will increase additional parameters $W$ of fully connected layers compared with the original softmax CE loss. 
Specifically, to prevent embedding models from learning noisy representations, Shi \etal proposed probabilistic face embeddings (PFEs \cite{Shi_2019_ICCV}).
But it needs additional calculations to estimate a distribution in the latent space. 
And the Arcface does not converge well with the PFEs. 
In contrast to those methods, ADR can be embedded into them easily without adding more cost.

\begin{figure*}[!htb]
	\begin{center}
		\includegraphics[width=\textwidth]{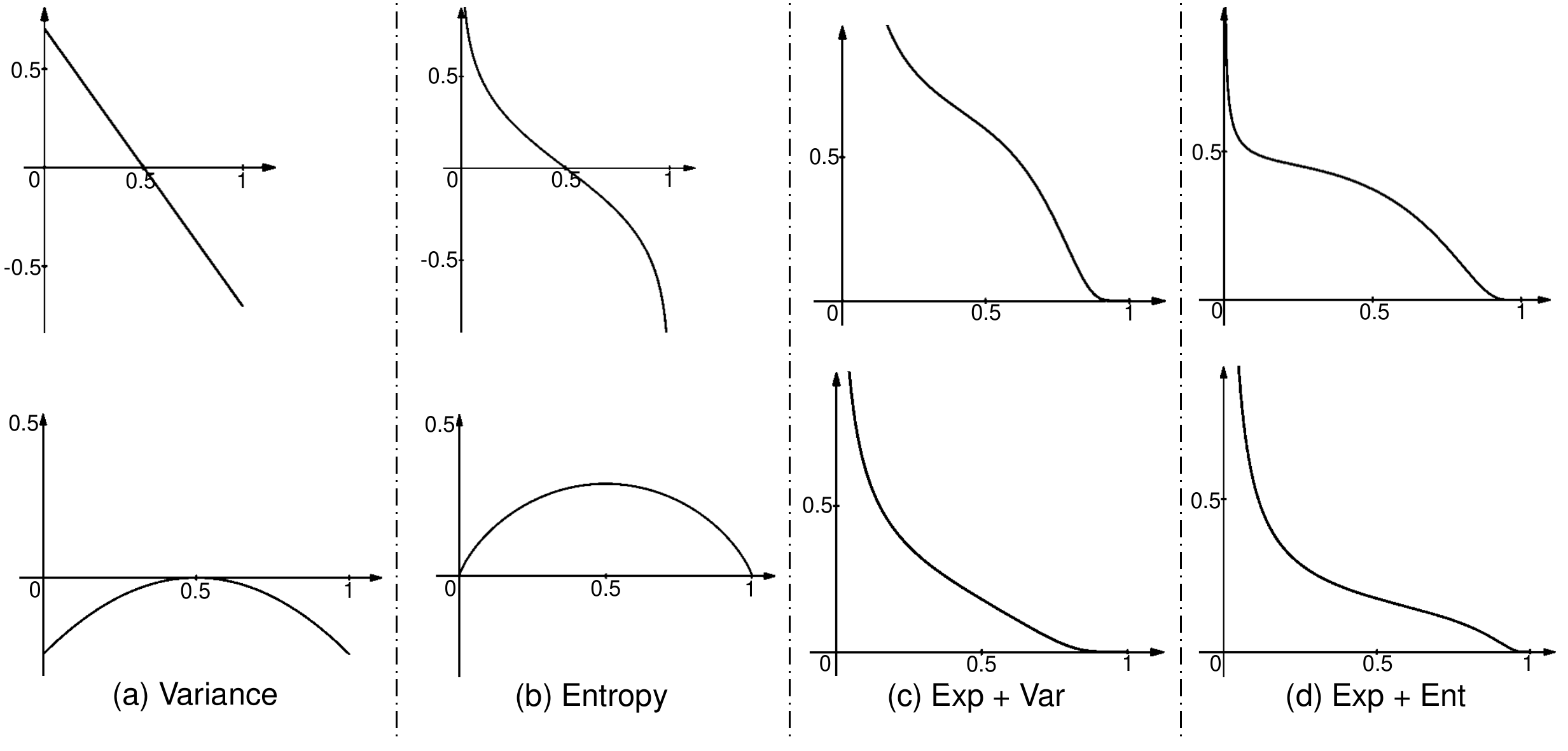}
	\end{center}
	\caption{ Visualization of the function curves. The discriminative function is shown below and its derivative function is shown above. (a) Variance-base functions. (b) Entropy-base functions (c) and (d) Motivation of our introduced ADR: (c) exponential-variance based functions, (d) exponential-entropy based functions in Eq. \ref{entropy_phi}. 
	}
   \vspace{-0.1in}
   \label{fig:curves}
\end{figure*}

\paragraph{Superset Label Learning. }
The superset label learning (aka partial label learning (\cite{wang2019partial,xu2021instance,wang2021adaptive})) is a machine learning paradigm that differs from conventional supervised learning, in which one training example can be ambiguously annotated with a set of labels, among which only one is correct. 
Existing methods for SLL commonly contain an explicit disambiguation operation to pick up the ground truth label of each training example from its candidate labels. 
For example, \cite{gong2017regularization} utilizes the $l^2_2$ norm (similar to the variance-base approach) as the discrimination term, and develops a regularization approach for the instance-based SLL.
\cite{yao2020deep} designs an entropy-base regularizer as the discrimination term to enhance the discrimination ability of the model.
In this paper, we try to formulate an adaptive discriminative regularization for supervised visual classification. 
Different from SLL, the supervised learning models need a larger backward gradient to enhance the confidence of predicted likelihoods at the beginning, while such a gradient should be small to avoid over-fitting when training is nearly ended.

\paragraph{Preliminaries. }
\label{subsuc:Preliminaries}
In the framework of maximum likelihood estimation (MLE), the cross entropy loss is employed for visual classification (\cite{de2005tutorial}). 
Suppose we have $n$ training instances $ {\cal X} = \{ x_1, x_2, \dots, x_n \}$ with the ground truth labels $ {\cal Y} = \{ y_1, y_2, \dots, y_n \}$, and every $y_i$  only has one explicit value $y_i \in \{1,2,\dots,c \} $, $c$ is the number of classes. 
In practice, we often divide the training data into $M$ batches $ {\cal X} = \{ {\cal X}_1, {\cal X}_2, \dots, {\cal X}_M \}$, and every ${\cal X}_m$ ($m=1,2,\dots,M$) contains $B$ samples.
After the feature embedding and the softmax function, every $x_i$ yields one prediction vector ${\tilde y}_i$. The ${\tilde y}_{ij}$ ($j=1,2,\dots,c$) denotes the predicted probability that the example $x_i$ belongs to the $j$-th category. 
The cross-entropy loss for a batch of images can be expressed as
\begin{equation}
\label{eq:ce_loss}
\begin{aligned}
{\cal L}_{ce}({\tilde {\cal Y}}_m, {\cal Y}_m) & = -\sum_{i=1}^{B} \sum_{j=1}^{c}  y_{ij} \text{log}({\tilde y}_{ij})) \\
\text{s.t. } \sum_{j=1}^{c} {\tilde y}_{ij} & = 1, {\tilde y}_{ij} \geq 0, \forall i = 1,2, \dots, B, \\ 
\end{aligned}
\end{equation}
For simplicity, we also give below the cross entropy and its derivative for the binary case.
\begin{equation}
	\label{eq:d_celoss}
	\left\{
	\begin{aligned}
		{\cal L}_{ce}(p_t,y) & = -log(p_t)\\
		\frac{\partial {\cal L}_{ce}}{\partial p_t} & = -\frac{1}{p_t},
	\end{aligned}
	\right.
\end{equation}
where $y \in \{ 0, 1 \} $ specifies the target, $p_t$ donates $p$ if $y=1$ or $1-p $ if $y=0 $ ($p \in \left [ 0, 1 \right ] $ ).
According to Eq. \ref{eq:d_celoss}, when the $p_t $ approaches $1$, the curve of cross entropy flattens out.

\section{The Proposed Method}\label{sec3}
\subsection{Intuition}
\label{Intuition}
In classical visual classification formulation, each visual sample (image or video) is assigned with a discrete label.
We then predict the likelihood about this sample belongs to the given classes and infer its estimated label is the one with the maximum logit.
Intuitively, the predicted logits should be sparse as only one class is indeed associated.
Optimizing the cross entropy between the predicted logits and the ground truth leads the predicted logits to the real data distribution.
With this premise, we suppose to encourage the predicted logits to be deterministic will benefit the whole optimization.

Intuitively, we could directly put sparsity regularization on the predicted logits using the approximated implementation,
e.g., minimizing $l_1$ norm of the differential between them, or their entropy.
Empirically, we find these solutions can accelerate the complete training but without evident performance increase. 
We attribute this to their coarse control of the regularization strength.
Specifically, the gradient of the $l_1$ norm is a constant $\mathbb{C} $, while the gradient of the entropy starts with zero and ends with larger values (as shown in Fig. \ref{fig:curves} (b)).
Neither of them is suitable for the optimization of supervised classification.
These observations and analysis inspire us to design an adaptive discriminative regularization. 
In detail, we expect the penalty on the logits is large when their distribution seems nearly uniform,
while it shrinks to small when their distribution changes to be spiky. 
Further, considering the optimization efficiency and stability, 
we prefer the change in the first-order derivative of this penalty to be huge when the predicted logits of each class are close to each other.
Meanwhile, such a change gets smaller when these logits differ from each other dramatically.

\subsection{Definition}
Based on our motivation, 
we present our adaptive discriminative regularization to improve visual classification optimization.
It automatically adjusts the discriminative constraints according to the confidence of the predicted logits.
Specifically, 
we hypothesize the penalties on the predicted likelihoods should be roughly linear to their own uncertainty as plotted in Fig. \ref{fig:curves} (d). 
\subsubsection{General form}
Our ADR for the supervised classification can be expressed as
\begin{equation}
	\label{general_form}
	\begin{aligned}
	{\cal L}_d({\tilde {\cal Y}}_m) & = \sum_{i=1}^{B} {\cal F}({{\tilde y}_i}), \\
	\text{s.t. } \begin{vmatrix} \frac{\partial {\cal F}({\tilde y}_i)}{\partial {\tilde y}_i} \end{vmatrix} & := 
	\left \{ \begin{aligned}
	& \approx  0^+, &  \phi({\tilde y}_i) \leq \epsilon  \\
	& \propto \phi({\tilde y}_i), & otherwise  \\
	\end{aligned} \right.
	\end{aligned}
\end{equation}
where ${\tilde y}_i \in {\tilde {\cal Y}}_m$,
and $\varphi_{i} =\phi({\tilde y}_i)$ measures the uncertainty of ${\tilde y}_i$ along with the non-negative and normalization constraints $ (0 < \varphi_{i} \leq 1)$. 
For example,
the entropy-base function satisfying this constraint, able to serve as this uncertainty function $\phi$.
$\epsilon$ is a threshold $(0 < \epsilon < 1)$ associated with $\varphi_{i} $ and $\tau $ ($\tau$ is a hyper-parameter in Eq. \ref{entropy_phi}).
${\cal F}(\cdot)$ donates the discriminative regularization function of our ADR.
As given in Eq. \ref{general_form},
we expect its partial gradient w.r.t. ${\tilde y}_i$ should be non-negative,
and is approximately proportional to the uncertainty of ${\tilde y}_i$ when ${\phi(\tilde y}_i)$ is larger than a given threshold $\epsilon$;
while it is smaller than $\epsilon$, such gradient is close to 0.

With this design of gradient changes,
it encourages the predicted logits to be certain and the overall training is adaptive based on the certainty of predictions. 
Specifically,
the optimization of visual classification along with this regularization method will be accelerated in the early training stage with the uncertain predicted logits
and will be accordingly slowed down when predictions tend to be deterministic. Ideally,
this regularization scheme leads to faster convergence and more stable performance during the training plateau.

\subsubsection{A simple solution}
\label{A Simple Solution}
An intuitive simple solution for our ADR (Eq. \ref{general_form}) is given below
\begin{equation}
    \label{entropy_phi}
    \begin{aligned}
     & {\cal F}({{\tilde y}_i}) = \frac{1}{(\sqrt{2\pi\varphi_{i}})^\tau} \text{exp} \{-\frac{1}{2\varphi_{i}}T({\tilde y}_{i})\}, \\
	& \text{s.t. } \left \{ \begin{aligned}	
	& \phi({\tilde y}_i) = -\frac{1}{B} \sum_{i=1}^{B} \frac{{\tilde y}_i^\top \text{log} ({\tilde y}_i)}{\text{log}(c)},  \\
	& T({\tilde y}_{i}) = \left\|\text{TopK}({\tilde y}_i, \tau) \right\|_2^2 ,\\
    \end{aligned} \right.
    \end{aligned}
\end{equation}
where $\varphi_{i}$ is a non-deterministic measure of the predicted vector. 
We apply the exponential function because of its natural simplicity properties.
Different from trigonometric and polynomial functions,
all the $n$-th order partial derivatives of the exponential are their own,
and they are always the same increasing functions.
Intuitively, 
the exponential form has better gradient diversity (see \cite{yin2018gradient}),
and that also can simplify the process of backward derivatives (turn to Sec. \ref{Backward Gradient} for details).
The other basis functions which are subject to Eq. \ref{general_form} also can be utilized to formulate the solution of our ADR. 

$T(\cdot)$ is called \textit{confidence-based normalization} function, characterizing the sufficient statistics (\cite{dynkin1978sufficient}) of the classes based on their predicted logits. $\text{TopK}(\cdot)$ is a non-linear sorting function. 
Utilizing sufficient statistics is to alleviate the negative influence brought by bad cases. 
For example, 
a classifier may misclassify tigers to cats in some scenarios with certain logits. Due to the appearance similarity between tiger and cat classes,
their logits will be close even tiger class has a higher logit.
With sufficient statistics, 
the transformation of the predicted logits will become more uncertain considering local class similarities,
giving proper gradient changes even with current wrong predictions.
Specifically, a logits vector ${\hat y}_i $ ($\sum_{j=1}^{\tau} {\hat y}_{ij} \leq 1$) of the ambiguous classes is selected by $\text{TopK}({\tilde y}_i, \tau)$ from the sparse ${\tilde y}_i$.
$\tau \in \mathbb{N}^+ $ is a hyper-parameter and indicates the assumed number of similar classes ($1 \leq \tau \leq c$ ).
The other non-linear functions which can pick out the local class similarities will work too.
For example, setting a confidence threshold to generate ${\hat y}_i $.
``$\left\| \cdot \right\|_2 $" computes the $l_2$ norm of the vector, 
and the $l_1$ norm ``$\left\| \cdot \right\|_1 $" may work too. 
As a result, a simple form of our approach to supervised classification can be written as
\begin{equation}
	\label{eq:final_loss}
	\begin{aligned}
		& \textit{Loss} = {\cal L}_{ce}({\tilde {\cal Y}}_m, {\cal Y}_m)  + \gamma {\cal L}_d({\tilde {\cal Y}}_m),
	\end{aligned}
\end{equation}
where $\gamma$ is one non-negative trade-off parameter controlling the relative weight of the ADR in the overall cost function.

\subsection{Discussion}
\label{Discussion}
This section further discusses the necessity of our ADR and clarifies the key operations in it.

\paragraph{Why does minimizing cross entropy need discriminative regularization?}
As we described in Sec. \ref{sec:Introduction}, 
classes in real-world datasets are not ideally independent of each other. 
They share high-level concepts more or less, 
e.g., cats and dogs both have four legs and fur, 
compared with airplanes or ships.
Simply put, 
we suppose that supervised tasks are subject to non-deterministic problems too,
a discriminative regularization favoring deterministic likelihoods could alleviate this issue. 
Specifically, 
we give a conceptual illustration of how cross-entropy and our ADR interact with each other from a toy example as given in Fig. \ref{fig:pipline}. 
This toy example is conducted on CIFAR-10 (\cite{krizhevsky2009learning}) with AlexNet (\cite{krizhevsky2014one}), and our ADR can improve the cross entropy performance up to 4.42\% (see Sec. \ref{sec:Image Classification}).
At different training phases,
we capture the predicted logits, and plot their change curve of ADR as given in Fig. \ref{fig:pipline} (middle).
In Fig. \ref{fig:pipline}, we observe that our adaptive discriminative regularization and orthodox cross entropy do \textbf{not} share the same optimization path in training.
The cross-entropy can be seen as the blue (upward) arrow in Fig. \ref{fig:pipline},
and its optimization direction is pulled to the positive side easily with the MLE algorithm. 
Given the right optimization direction,
we suppose our given ADR can further pull the optimization toward one end along the (yellow) line where the predicted logits are located. In Fig. \ref{fig:pipline}, we find
the predicted likelihoods are amplified with ADR, making the optimization quickly fall to one side further.
Hence, discriminative regularization is not only suitable for supervised tasks, but also requires fine-grained design.

\paragraph{How about employing entropy as a discrimination term?}
We first overview the stochastic gradient descent (SGD) algorithm (\cite{qian1999momentum}).
It follows the estimated gradient downhill as
\begin{equation}
	\label{sgd}
	\left\{
	\begin{aligned}
		& {\cal G} = \frac{\partial {\cal L}(\theta)}{\partial \theta}\\
		& \theta \leftarrow  \theta - \alpha {\cal G}, \\
	\end{aligned}
	\right.
\end{equation}
where $\theta $ denotes the learnable parameters,
$\alpha $ is the learning rate,
${\cal G} $ is the estimated gradient descent of the cost functions.
The SGD updates $\theta$ by calculating the partial derivatives of the cost function at each parameter, i.e., $\nabla_\theta {\cal L}(\theta) $,
and it often finds a low value of the cost function quickly. 
We find that the learning step of every update of $\theta$ is \textbf{not} proportional to the value the loss function takes,
but the partial derivative it makes.
From this insight,
the existing discrimination terms only pay attention to the functionality of the regularization constraints,
i.e., the largest ${\cal L}(\theta)$ corresponds to the most ambiguous $\theta$, and vice versa. 
However, we advocate that in addition to the above criteria,
the timing and magnitude of the regularization intervention should be taken into account specifically.

Different discriminative regularizations are drawn in Fig. \ref{fig:curves}.
Observing the derivative function curves,
we find that different from the entropy function, 
the proposed exponential-based solutions yield a larger gradient for the early to mid-stage of training.
Additionally, 
to prevent over-confident predictions the gradient of our solution rapidly closes to zero later in training.
Utilizing the entropy as a discrimination term (e.g., \cite{li2003efficient, yao2020deep}) to widen the gap of predicted likelihoods could work well in SLL. 
Because the SLL aims to the problem that a training example is associated with a set of candidate labels.
In detail, the gradient of entropy could be zero as we do not know whether the optimization direction is right at the beginning, and it holds a large value due to the model already knows where the positive side is later in training.
Therefore,
employing entropy as the discriminative regularization in Eq. \ref{eq:final_loss} will lead to severe optimization issues theoretically and empirically,
while the given exponential-based solutions will not.

\paragraph{For the uncertainty function $\phi$ in Eq. \ref{entropy_phi}, entropy-base better than variance-base?}
The variance describes the variation of one random variable while the entropy represents the uncertainty of the information, both of them can be used as an uncertainty function $\phi$ which is designed to evaluate the confidence of predicted likelihoods.
However,
when a random variable obeys a non-convex distribution, 
the ability of the variance to describe the information uncertainty will reduce,
while the entropy could do better (turn to \cite{zidek2003uncertainty} for details).
As shown in Fig. \ref{fig:curves}, we give two specific implementations for it,
in which the derivative of the exponential variance decreases faster than that of the exponential entropy. 
We suppose that the effective interval of its derivative function is then small.
Hence, we pick the latter (the exponential entropy in Fig. \ref{fig:curves} (d)) as the default solution of our ADR throughout this paper.

\paragraph{Gradient of ADR. }
\label{Backward Gradient}
Not only does the ADR suit our requirements on its changes, but also it is easy to be optimized according to its gradient form.  
According to Eq. \ref{entropy_phi}, 
the solution of our ADR for a single sample ${\cal L}_d({\tilde y}_i)$ can be rewritten as
\begin{equation}
    \label{ADR_sample}
    \begin{aligned}
    {\cal L}_d({\tilde y}_i) & = \prod_{j=1}^{\tau} \frac{1}{\sqrt{2\pi\varphi_{i}}} \text{exp} \{-\frac{1}{2\varphi_{i}}{\hat y}^2_{ij}\}. \\
    \end{aligned}
\end{equation}
We only calculate the partial derivative $ \frac{\partial {\cal L}_{d}({{\tilde y}_i})}{\partial {\tilde y}_{i}} $ for simplicity, and it can be computed via
\begin{equation}
    \label{simple_partial}
    \begin{aligned}
    \frac{\partial {\cal L}_{d}({{\tilde y}_i})}{\partial {\tilde y}_{i}} & = \sum_{j=1}^{\tau} \left[ {\cal L}_{d}({{\tilde y}_i}) \cdot \frac{{\hat y}_{ij}^2 \varphi'_{ij} - 2{\hat y}_{ij}\varphi_{i} - \varphi_{i}\varphi'_{ij} }{2\varphi_{i}^2} \right],  \\ 
    \text{s.t. } \varphi'_{ij} & = \frac{\partial \varphi_{i}}{\partial {\hat y}_{ij}}.
    \end{aligned}
\end{equation}

The backward derivation (above equation) contains the results of the forward propagation calculation (e.g. Eq. \ref{ADR_sample}), 
which will reduce the time complexity of our ADR significantly.
Specifically, the naive computation of Eq. \ref{simple_partial} requires only $O(\tau^2)$ operations, 
as the terms $\varphi_{i}$ and ${\cal L}_{d}({{\tilde y}_i}) $ can be computed once and reused in each derivation.

\begin{table}[!tb]
\setlength\tabcolsep{2pt}
\begin{center}
\caption{Recognition Top-1 error rate on ImageNet-1K classification benchmark.}\label{table:ilsvr2012}
\resizebox{\linewidth}{!}{
\begin{threeparttable}
\begin{tabular}{lllcl}
\toprule
Model &Param &Method  & Epochs  &  Top-$1$\%   \\ 
\toprule
RN-50  &25.6M  &CE loss       &250    &23.68\dag         \\ 
RN-50  &25.6M   &LS (\cite{szegedy2016rethinking})  &250    &22.82\dag       \\ 
RN-50  &25.6M   &CutOut (\cite{devries2017improved})  &250   &22.93\dag  \\ 
RN-50  &25.6M   &BYOT (\cite{zhang2019your})   &250    &23.04\dag    \\ 
RN-50  &25.6M   &$Tf$-$KD_{self}$ (\cite{yuan2020revisiting})  &90     &23.59\dag  \\ 
RN-50  &25.6M   &$Tf$-$KD_{reg}$ (\cite{yuan2020revisiting})  &90     &23.58\dag   \\ 
RN-50  &25.6M$^+$   &OLS (\cite{zhang2021delving}) &250    &22.28\dag      \\
RN-101 &44.7M  &CE loss            &250    &21.87\dag           \\
RN-101 &44.7M    &LS (\cite{szegedy2016rethinking})  &250    &21.27\dag     \\ 
RN-101 &44.7M    &CutOut (\cite{devries2017improved})  &250    &20.72\dag   \\ 
RN-101 &44.7M$^+$  &OLS (\cite{zhang2021delving})  &250    &20.85\dag      \\
I-V2 &23.9M  &CE loss  & - &23.10$\ddag$  \\
I-V2 &23.9M  &LS (\cite{szegedy2016rethinking})  & - &22.80$\ddag$  \\
I-V4 &43.0M  &CE loss & - &19.10$\ddag$  \\
I-V4 &43.0M  &LS (\cite{szegedy2016rethinking})  & - &19.10$\ddag$ \\
\toprule
RN-50  &25.6M  &CE loss   &100       &23.06   \\  
RN-50  &25.6M  &w/ ADR      &100   &22.49  \\  
RN-101 &44.7M   &CE loss  &100      &21.30    \\
RN-101 &44.7M   &w/ ADR      &100    &20.76   \\ 
ViT-B/16 &86.6M  &LS      & 300 &18.05   \\
ViT-B/16 &86.6M  &w/ ADR     & 300   &\textbf{17.69}     \\ 
\toprule
\end{tabular}
\begin{tablenotes}
    \item \dag and $\ddag$ denote the results reported in \cite{zhang2021delving} and \cite{muller2019does} respectively. $^{+}$ means the addition of some parameters.
    \item ``ResNet" is abbreviated as ``RN", and ``I-V2" means ``INCEPTION-V2". 
\end{tablenotes}
\end{threeparttable}}
\end{center}
\end{table}

\section{Experiments}
\label{Experiments}
To evaluate the proposed ADR,
we conduct extensive experiments on five typical vision applications,
including image classification (ImageNet-1K (\cite{russakovsky2015imagenet}), Flowers-102 (\cite{Nilsback08}), and CIFAR-10),
face verification (CASIA (\cite{yi2014learning}), etc.),
facial emotion recognition (FER2013 (\cite{goodfellow2013challenges})),
action recognition (NTU RGB+D \cite{shahroudy2016ntu}),
and unsupervised image segmentation (PASCAL VOC 2012 (\cite{everingham2015pascal}) BSDS500 (\cite{arbelaez2010contour})).

\begin{table}[!tb]
\begin{center}
\caption{Recognition accuracy on Flowers-102 with the architecture of ResNet-50. }\label{table:Flowers}
\setlength\tabcolsep{6pt}
\resizebox{1.0\linewidth}{!}{
\begin{threeparttable}
\begin{tabular}{lllll}
\toprule
Model  &Method  &Pub.'Year  &Top-1\%  &Top-5\% \\ 
\toprule
RN-50       & CE loss     &TIP'21   &90.69\dag                       &97.57\dag \\  
RN-50       & CE+LS       &TIP'21   &92.42\dag                       &98.07\dag \\
RN-50       & CE+OLS      &TIP'21   &92.86\dag                       &98.45\dag \\
\toprule
RN-50       & CE loss     &TIP'21   &90.89                           &97.45 \\  
RN-50       & CE+LS       &TIP'21   &92.12                           &97.71 \\
RN-50       & CE+OLS      &TIP'21   &93.12                           &98.31 \\
RN-50       & CE+ADR      &-     &92.36 ($1.47\uparrow$) &98.10 \\
RN-50       & LS+ADR     &-     &93.12 ($1.00\uparrow$) &98.28 \\
RN-50       & OLS+ADR     &-     &\textbf{93.49} ($0.37\uparrow$) &98.05\\
\toprule
\end{tabular}
\begin{tablenotes}
    \item \dag denotes the results reported in \protect\cite{zhang2021delving}. ResNet is abbreviated as RN. 
\end{tablenotes}
\end{threeparttable}}
\end{center}
\end{table}

\begin{table}[!tb]
\caption{Recognition accuracy of different methods on CIFAR-10. }\label{table:cifar10}
\begin{center}
\setlength\tabcolsep{5pt}
\resizebox{1.0\linewidth}{!}{
\begin{threeparttable}
\begin{tabular}{lllll}
\toprule
Model  &Method  &Pub.'Year  &Top-1\%  &Top-5\% \\ 
\toprule
AlexNet\dag    &CE loss     &NIPS'19   &86.80\dag          &- \\ 
AlexNet\dag    &LS           &NIPS'19  &86.70\dag           &- \\ 
\toprule
AlexNet    &CE loss      &arXiv'14        &73.33                           &98.01 \\ 
AlexNet    &CE+Entropy   &-        &74.21 ($0.88\uparrow$)          &98.02 \\ 
AlexNet    &CE+ADR       &-     &77.07 ($3.74\uparrow$)          &98.00 \\
AlexNet    &LS           &CVPR'16  &76.31 ($2.98\uparrow$)          &98.09 \\ 
AlexNet    &LS+ADR       &-     &77.75 ($4.42\uparrow$) &97.60 \\
ConvMixer &CE loss      &arXiv'22 &89.79                           &99.70 \\  
ConvMixer &CE+ADR         &-     &\textbf{92.47} ($2.68\uparrow$) &\textbf{99.78} \\
\toprule
\end{tabular}
\begin{tablenotes}
    \item \dag denotes the model and results are described in \cite{muller2019does}.
\end{tablenotes}
\end{threeparttable}}
\end{center}
\end{table}

\begin{table}[!tb]
\caption{The top-1 accuracy comparison of CE and ADR on CIFAR-10 with adding different noise rates (NR). }\label{table:noise}
\begin{center}
\setlength\tabcolsep{3pt}
\resizebox{1.0\linewidth}{!}{
\begin{tabular}{l|llll}
\toprule
Method  & NR=20\%  & NR=40\%  & NR=60\%  & NR=80\%\\ 
\toprule
CE loss   &72.51        &68.95       &63.45        &43.38\\  
CE+ADR      &\textbf{74.89} ($2.38\uparrow$) &\textbf{70.53} ($1.58\uparrow$)  &\textbf{63.65} ($0.20\uparrow$)    &\textbf{43.83} ($0.45\uparrow$)\\
\toprule
\end{tabular}}
\end{center}
\end{table}

\paragraph{Experimental Settings.} 
In all the experiments,
we use the same neural network architecture and experimental environment (Pytorch 1.7.0 on NVIDIA 1080Ti) for fair comparisons.
Different losses are employed to outline the properties of ADR.
The applied neural models include such CNN-based ones with different depths and structures as AlexNet (\cite{krizhevsky2014one}), VGGNet (\cite{khaireddin2021facial}), ResNet-50/101 (\cite{he2016deep}) and the extended ResNet3D-34 (\cite{ji2021exploiting}). 
Transformer based architectures are evaluated, e.g., ViT (\cite{dosovitskiy2020vit}) and ConvMixer (\cite{trockman2022patches}), as well.

\subsection{Image Classification}
\label{sec:Image Classification}
\paragraph{ImageNet-1K.} 
We employ ResNet-50/101 in \cite{Radosavovic2020} as backbones and perform all experiments by utilizing the same training/testing protocols as in \cite{szegedy2016rethinking} and \cite{Radosavovic2020}.
We also use a vision transformer architecture ViT-B/16 (\cite{dosovitskiy2020vit}) as the backbone and follow the DeiT (\cite{v139touvron21a}) training configuration for training.
The results are reported in Table \ref{table:ilsvr2012}. 
Our ADR consistently decreases the error rate on ImageNet-1K with ResNet-50/101 more than $0.5\%$ ($0.57\%$ with ResNet-50 and $0.54\%$ with ResNet-101). 
That validates the effectiveness of the proposed ADR on a large-scale supervised classification dataset with a decently large convolution-based model.
Also, such effectiveness is further verified with a popular vision transformer model (ViT-B/16) by reducing the error rate $0.36\%$.
It shows the performance improvement brought by ADR is relatively agnostic to model architecture.
\paragraph{Flowers-102. }
We find ADR works fine for fine-grained discriminative tasks.
We conducted experiments by following the same architecture of ResNet-50 as \cite{zhang2021delving}.
The results are given in Table \ref{table:Flowers}.
They demonstrate that our ADR can handle fine-grained classification.
Based on the standard CE, LS, and OLS, the additional ADR can achieve a notable improvement by 1.47\%, 1.00\%, and 0.37\%, respectively.
\begin{figure}[!tb]
\begin{center}
\includegraphics[width=1.0\linewidth]{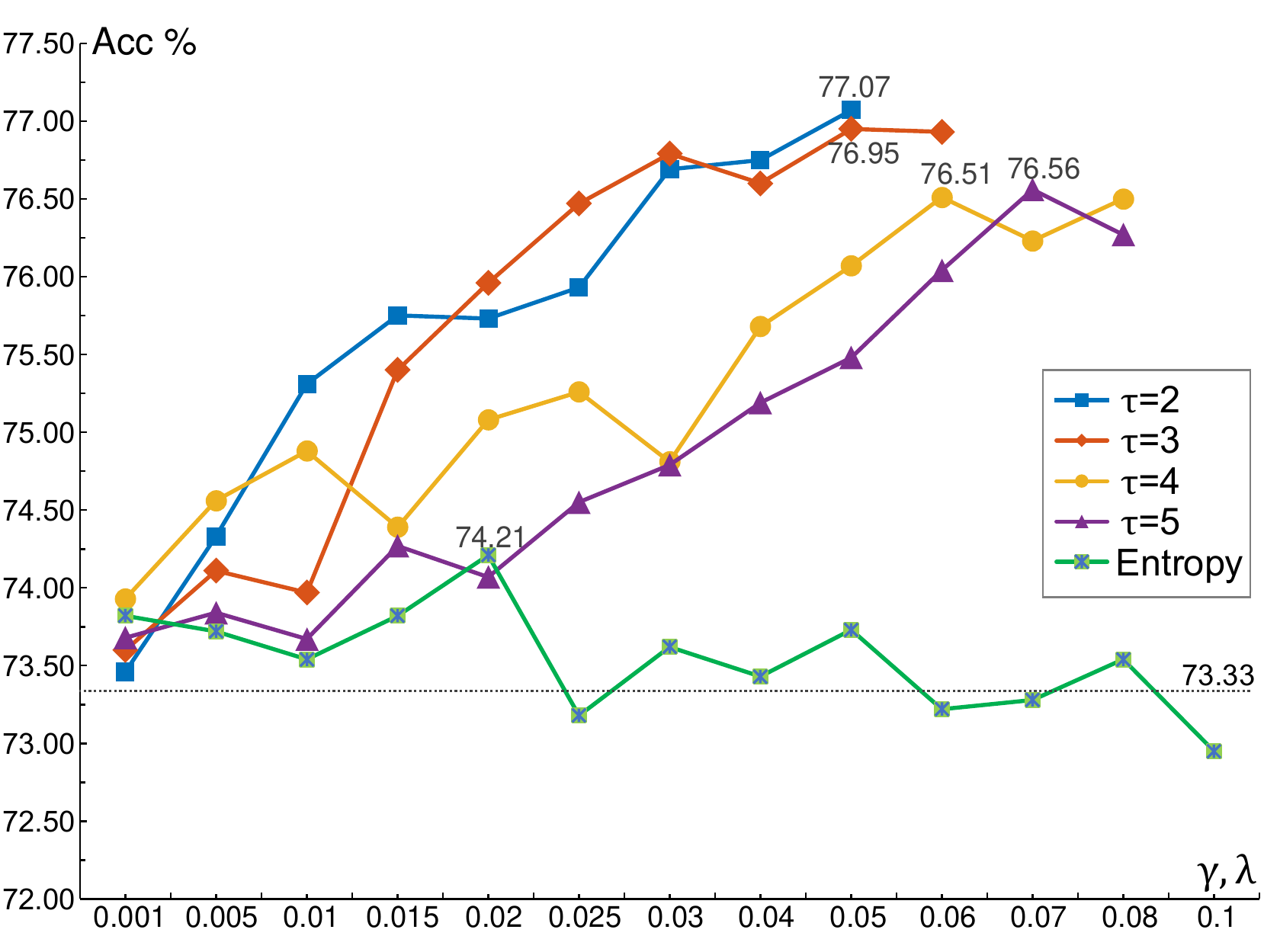}
\end{center}
\caption{The classification performance of ADR and the entropy loss, varying $\gamma,\lambda$ for ADR (w. optimal $\tau$) and the entropy loss respectively.}
\label{fig:varying_gamma}
\end{figure}

\begin{table}[!tb]
\caption{Face datasets for training and validation.}\label{table:face_datasets}
\begin{center}
\setlength\tabcolsep{5pt}
\resizebox{1.0\linewidth}{!}{
\begin{tabular}{lll}
\toprule
Datasets & \#Identity  & \#Image \\ 
\toprule
LFW (\cite{huang2008labeled}) & 5749     &   13233 \\ 
CASIA (\cite{yi2014learning}) & 10K      &   0.5M \\ 
CFP-FP/FF (\cite{sengupta2016frontal}) & 500     &   7000 \\ 
AgeDB-30 (\cite{moschoglou2017agedb})  & 568     &   16488 \\ 
CALFW (\cite{zheng2017cross}) & 5749     &   12174 \\
CPLFW (\cite{zheng2018cross}) & 5749     &   11652 \\ 
\toprule
\end{tabular}}
\end{center}
\end{table}

\paragraph{CIFAR-10.} 
\label{CIFAR10}
Also, we employ the neutered version of AlexNet (without batch normalization layers \cite{krizhevsky2014one}) and one advanced Transformer-based model ConvMixer as the backbones.
We refer to the commonly used protocols with data augmentation in \cite{lee2015deeply} for training.
Quantitative results are shown in Table \ref{table:cifar10}. 
Compared to the CE, Entropy, and LS, the proposed ADR obtained the best performance boost on both very different backbone baselines (i.e. $3.74\%$ with AlexNet and $2.68\%$ with ConvMixer).

\begin{figure*}[!tb]
\begin{center}
\includegraphics[width=0.93\linewidth]{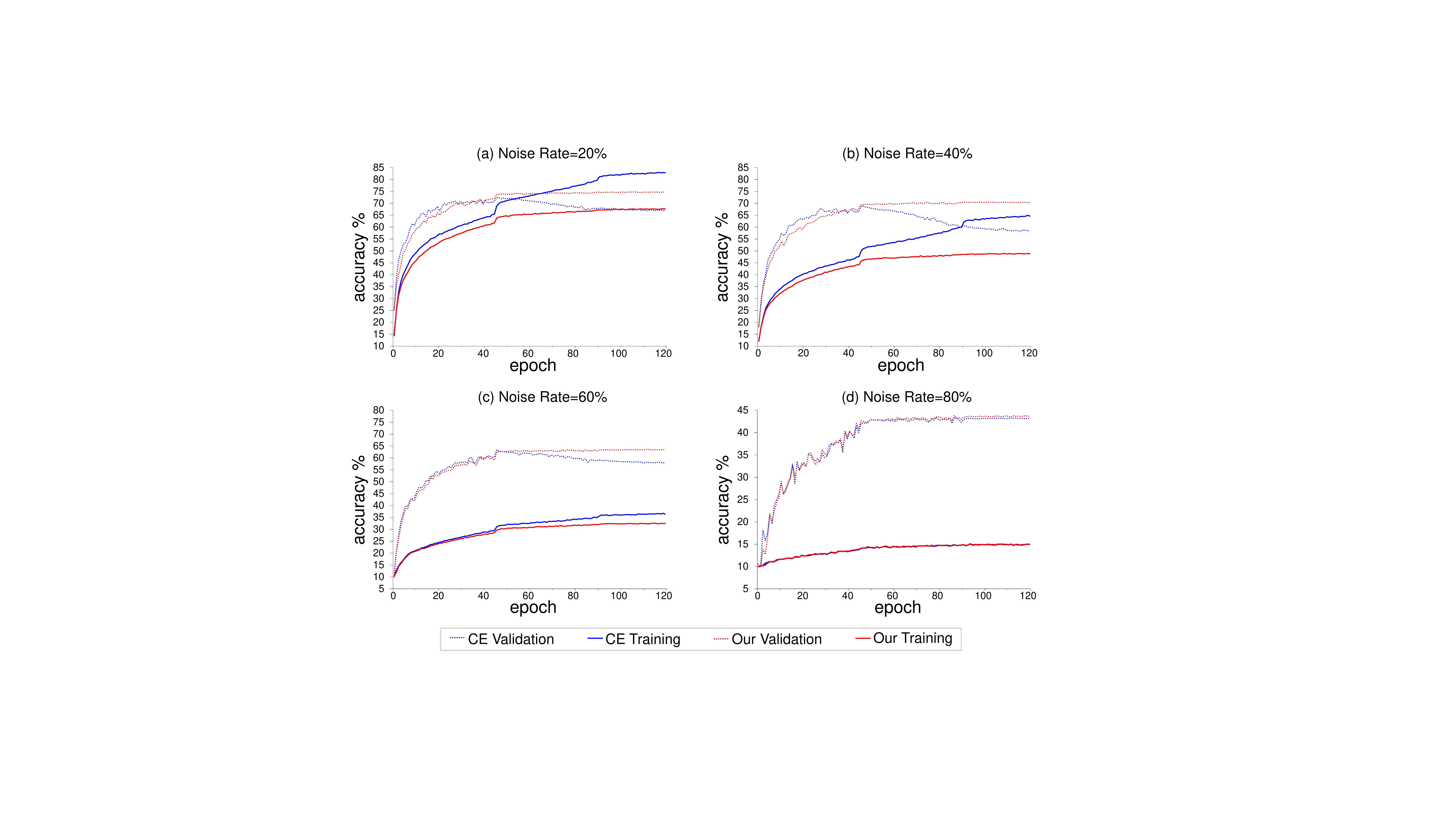}
\end{center}
\caption{Classification accuracy under different noise rates.
The accuracy of ADR can increase rapidly and eventually become steady with the different noisy labels}
\label{fig:noise}
\end{figure*}

The entropy loss can be embedded in the cross entropy too,
we conduct experiments to compare our ADR against it. 
Following the entropy formulation defined in \cite{yao2020deep},
we define the total loss function as:
\begin{equation}
\label{eq:entropy loss}
\textit{Loss} ={\cal L}_{ce}({\tilde {\cal Y}}_m, {\cal Y}_m)  + \lambda {\cal L}_e({\tilde {\cal Y}}_m), 
\end{equation}
where $\lambda$ denote the strength of the modulating term of ${\cal L}_e({\tilde {\cal Y}}_m)$. 
As shown in Table \ref{table:cifar10},
there is a clear gap between the best accuracy of the entropy (74.21\%) and ADR (77.07\%).

\subsubsection{Ablation Study}
Specifically, we investigate the influences of hyper-parameters and tolerance of ADR to noisy labels on CIFAR-10.

\paragraph{Sensitivity to Parameters.}
The tradeoff parameter $\gamma$ is to control the strength of the proposed ADR.
With larger $\gamma$, the discriminative force in Fig. \ref{fig:pipline} becomes larger and the predicted logits become sparser.
Another parameter $\tau$ closely determines the value of the threshold $\epsilon $ in Eq. \ref{general_form}.
With smaller $\tau$, the discriminative force will increase and the strength of ADR $\gamma$ should be smaller accordingly.
The relation between the accuracy and $\gamma$ with different $\tau$ on CIFAR-10 is reported in Fig. \ref{fig:varying_gamma}. 
With the increase of $\gamma$ and a fixed $\tau$, one can see that the accuracy of ADR is always on the rise.
Still and all, the performance of those parameter settings outperforms the baseline $73.33\%$ (the gray dotted line) by at least $0.13\%$.
We observe that proposed ADR can achieve better performance with $0.01 < \gamma < 0.1 $, and the empirical value $\tau$ is proportional to the number of classes $\tau \propto c $.
Furthermore, the effectiveness of parameter $\lambda$ (the green one) for the Entropy loss is also drawn in Fig. \ref{fig:varying_gamma}.
We see that the accuracy-$\lambda$ curve of the Entropy is jittering up and down around the baseline.

\paragraph{Tolerance to Noisy Labels.}
We present how ADR reacts to noisy labels in classification. Generally, training with ADR will prevent the model to overfit noisy distribution.
We conducted the experiments using the same settings as \cite{wang2019symmetric}.
A certain number of samples are randomly selected and flipped to the uncorrected labels before training.
The Top-1 recognition accuracy results under four noisy rates  (20\%, 40\%, 60\%, 80\%) are reported in Table \ref{table:noise}.
In addition, the training and test accuracy vs. iteration is visualized in Fig. \ref{fig:noise}.
When an added noise rate is less than 50\%, our ADR could obtain a more stable improvement than the CE loss.
Note when noise rates exceed 20\% or 40\%, CE causes training overfitting as the training and test accuracy curves are intersected, and its training accuracy exceeds 80\% and 60\% respectively.
In contrast, the proposed ADR not only enables the model to be trained robustly, but also achieves a better classification performance.

However, when adding a high noise rate ($NR > 50\%$), the model with CE loss would be under-fitted as more than half training samples are useless or distracting. ADR also fails with heavy noise. After all, in heavy noise conditions, the calibration step (confidence-based normalization) in ADR is meaningless as there barely exists reliable semantically similar classes.

\begin{table*}[!tb]
\begin{center}
\setlength\tabcolsep{8pt}
\caption{Face Verification performance of different methods on LFW, CALFW, CPLFW, AgeDB-30, CFP-FP and CFP-FF datasets with ResNet18.}\label{table:face_result01}
\resizebox{1.0\linewidth}{!}{
\begin{threeparttable}
\begin{tabular}{lllllllll}
\toprule
Method  &Pub.'Year  & LFW\% & CALFW\%  &CPLFW\% &AgeDB-30\%  & CFP-FF\% &CFP-FP\%  & Average\% \\ 
\toprule
RBM (\cite{cao2020domain}) &CVPR'20 &99.10\ddag  &91.00\ddag  &87.10\ddag  &91.30\ddag  &-  &-  &92.125 \\
DBM (\cite{cao2020domain}) &CVPR'20 &99.20\ddag  &92.00\ddag  &87.30\ddag  &91.90\ddag  &-  &-  &92.600 \\
R\&D-BM (\cite{cao2020domain}) &CVPR'20 &\textbf{99.30}\ddag  &92.50\ddag  &\textbf{87.60}\ddag  &92.10\ddag  &-  &-  &92.875 \\
Arcface (\cite{deng2019arcface})   &CVPR'19 &99.10$\ast$ &89.05$\ast$ &78.43$\ast$ &93.18$\ast$ &-  &-  &89.940 \\
MFR (\cite{guo2020learning}) &CVPR'20 &99.12$\ast$ &89.45$\ast$ &79.22$\ast$ &93.30$\ast$ &-  &-  &90.273 \\
TigthROI (\cite{xu2021searching}) &AAAI'21 &99.02$\ast$ &88.78$\ast$ &79.30$\ast$ &93.73$\ast$ &-  &-  &90.208 \\
SuperROI (\cite{xu2021searching}) &AAAI'21 &99.18$\ast$ &88.80$\ast$ &79.22$\ast$ &93.38$\ast$ &-  &-  &90.145 \\
$\text{FAPS}_C$ (\cite{xu2021searching}) &AAAI'21 &99.20$\ast$ &89.47$\ast$ &80.28$\ast$ &\textbf{94.02}$\ast$ &-  &-  &90.743 \\
\toprule
Cosface (\cite{wang2018cosface}) &CVPR'18 &99.100          &93.033 &86.783 &93.167 &99.429          &92.871  &93.021 \\ 
w/ ADR    &-    &\textbf{99.300} &93.267 &87.483 &93.750 &\textbf{99.529} &93.057  &\textbf{93.450} \\
Arcface (\cite{deng2019arcface}) &CVPR'19 &99.200 &93.300          &86.833 &93.267 &99.414          &93.343          &93.150 \\  
w/ ADR    &-    &99.283 &\textbf{93.417} &87.050 &93.950 &\textbf{99.529} &\textbf{93.557} &93.425 \\
\toprule
\end{tabular}
\begin{tablenotes}
\item $\ddag$ and $\ast$ denote the results reported in \cite{cao2020domain} and \cite{xu2021searching}, respectively.
\end{tablenotes}
\end{threeparttable}}
\end{center}
\end{table*}

\begin{table}[!tb]
\setlength\tabcolsep{5pt}
\begin{center}
\caption{Recognition accuracy of test and validation data on FER2013 dataset. }\label{table:Fer2013}
\resizebox{0.75\linewidth}{!}{
\begin{tabular}{lll}
\toprule
Method   & Test \%    & Validation \% \\ 
\toprule
CE loss     &72.12        &73.82           \\  
CE+ADR        &\textbf{72.72} ($0.60\uparrow$)  &\textbf{74.39} ($0.57\uparrow$)  \\
\toprule
\end{tabular}}
\end{center}
\end{table}

\subsection{Face Verification}
ADR works fine in face verification. 
We utilize the same settings in Arcface for the following experiments. 
The ADR was embedded into the architecture of ResNet18. 
For model training, we employ the commonly used web-collected outside dataset CASIA (\cite{yi2014learning}) (excluding the images of identities appearing in the test set) which has $\sim 0.5M$ face images belonging to $\sim 10K$ different individuals.

For the validation, six datasets including LFW, CALFW, CFP-FF, CPLFW, CFP-FP, and AgeDb-30 are utilized to evaluate the performance. LFW includes $\sim 13K$ web-collected images from $\sim 5K$ different identities, with limited variations in pose, age, expression, and illuminations.
CPLFW was collected from LFW with a larger pose gap. 
Similar to CPLFW, CALFW was selected from LFW with higher variations of age. 
CFP consists of collected images of celebrities in frontal and profile views, which has two evaluation protocols consisting of CFP-Frontal-Frontal and CFP-Frontal-Profile which is a more challenging protocol with around a $90^\circ$ pose gap within positive pairs.
AgeDB-30, a ``in-the-wild" dataset, contains manually annotated images. In this paper, we employ the evaluation protocol with a 30-year gap.
Table \ref{table:face_datasets} lists the details of these datasets.

\begin{table}[!tb]
\caption{Recognition accuracy of cross-subject and cross-view evaluations on NTU RGB+D.}\label{table:ntu rgbd}
\begin{center}
\setlength\tabcolsep{5pt}
\resizebox{0.75\linewidth}{!}{
\begin{tabular}{lll}
\toprule
Method    & X-Sub \%  & X-View \% \\ 
\toprule
CE loss    &87.87    &90.82    \\  
CE+ADR      &\textbf{88.56} ($0.69\uparrow$) &\textbf{91.37} ($0.55\uparrow$)  \\
\toprule
\end{tabular}}
\end{center}
\end{table}

\subsubsection{Ablation Study}
Table \ref{table:face_result01} presents the results of ADR on such common datasets as LFW, CALFW, CFP-FF, CPLFW, CFP-FP and AgeDb-30.
For LFW and CFP-FP, ADR can boost the accuracy with any $\gamma,\tau$ settings,
raising by 0.017\%-0.214\% compared with both baselines. 
One can see that ADR can boost the performance over the baselines on CALFW ($0.117\%$ and $0.234\%$ respectively). 
And ADR can further reduce the error rates from $\approx 0.6\%$ to $\approx 0.5\%$ on CFP-FF. 
Specifically, ADR can outperform both the baselines by obvious margins ($0.683\%$ on AgeDb-30 and $0.700\%$ on the challenging CPLFW respectively).

\begin{figure*}[!tb]
\begin{center}
\vspace{0.1in}
\includegraphics[width=0.93\linewidth]{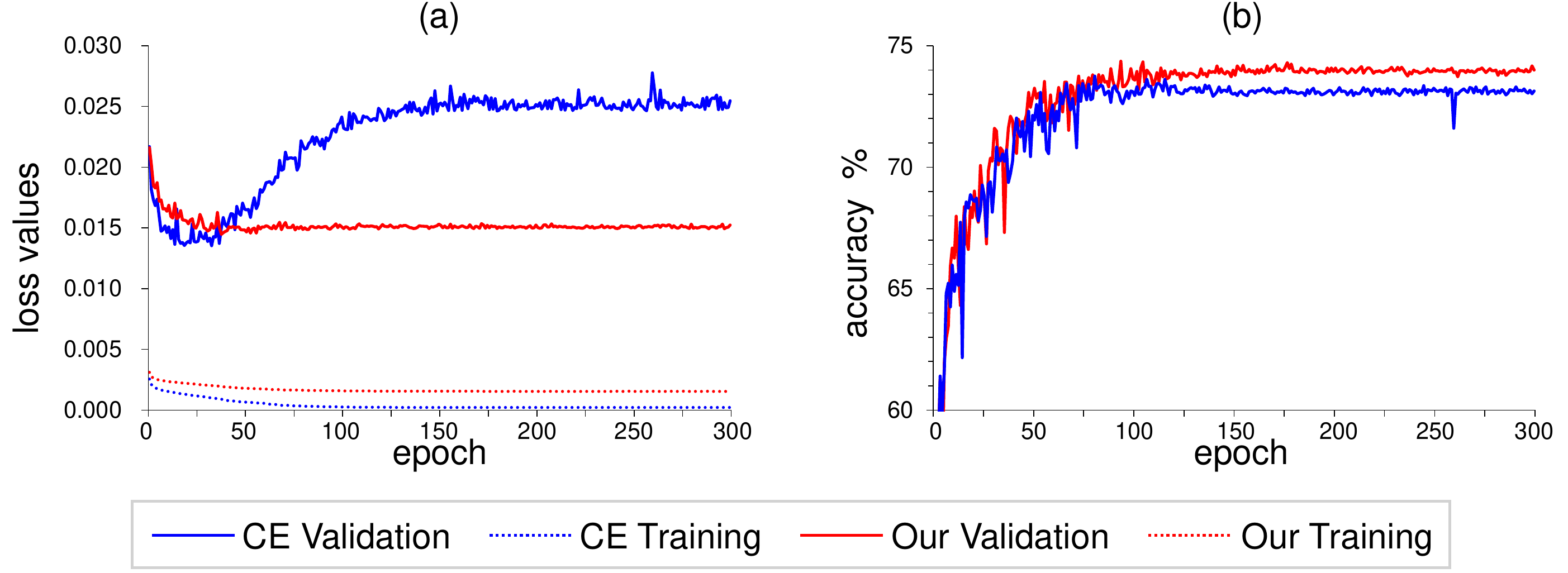}
\end{center}
  \caption{Recognition loss values and accuracy vs. epoch on FER2013 with different methods. See (a), the loss value of ADR decreases rapidly during the first 50 epochs and then gets convergence gradually.}
\label{fig:fer2013_loss_acc}
\end{figure*}

\subsubsection{Comparison with SOTA}
We evaluate ADR on serveral state-of-the-art (SoTA) face verification methods, including TigthROI (\cite{xu2021searching}), (R+D)BM (\cite{cao2020domain}) and $\text{FAPS}_C$ (\cite{xu2021searching}) etc. 
They are introduced in Table \ref{table:face_result01}. 
The performance of our ADR is superior to all other losses on LFW, CALFW, CFP-FF/FP, and the average accuracy of those 6 datasets. 
ADR can further improve current SOTA face verification results notably.

\renewcommand{\thefootnote}{\arabic{footnote}}

\subsection{Facial Emotion Recognition}
\label{subsec:Facial Emotion Recognition}
The \textbf{long-tailed} FER2013 dataset contains $\sim 36K$ images.
It has $7$ emotion classes, i.e., anger, neutral, disgust, fear, happiness, sadness, and surprise. 
We employ a customized VGGNet (\cite{khaireddin2021facial}) with an SGD optimizer to conduct experiments, following the official protocols in \cite{goodfellow2013challenges}.
In practice, we use the ReduceLROnPlateau\footnote{Pytorch 1.9.0 documentation: \url{https://pytorch.org/docs/stable/_modules/torch/optim/lr_scheduler.html}} against other schedulers to obtain a robust baseline as reported in Table \ref{table:Fer2013}.
Our ADR is superior to both the baselines of the test and validation sets by $0.6\%$ and $0.57\%$ respectively.

As shown in Fig. \ref{fig:fer2013_loss_acc}(a), we observe that the model tuned with the vanilla CE overfits the training data, as the training loss continues to decline, while the validation loss is increasing.
In our option, to further reduce the training loss, CE focuses on fitting well-classified samples.
That is, the score of class label $y=1$ in the validation set is reduced but still gives the correct prediction results. 
In contrast, the proposed ADR continues to decline until being stable in both training and validation sets. 
Note that the training loss of the ADR is larger than CE, but the validation accuracy is greater than it. 
It empirically validates our ADR focuses on fitting the hard samples (which are misclassified early) and can alleviate the overfitting in training.

\subsection{Action Recognition}
To verify the effectiveness of the proposed ADR for the sequence-based action recognition (\cite{kong2022human}), we employed the same experiment settings of ResNet3D-34 (\cite{ji2021exploiting}) (a Spatio-temporal network). 
The SGD optimizer with fixed momentum of $0.9$ and weight decay of $10^{-5}$ was utilized for training. 
We set the LR at $0.01$, and it decreases to one-tenth times every 20 epochs.

We apply the NTU RGB+D dataset as a benchmark, which consists of $60$ classes and contains $56880$ action sequences captured with three cameras from different views.
And we follow the cross-subject (X-Sub) and cross-view  (X-View) protocols introduced by \cite{shahroudy2016ntu} to conduct all experiments. 
The results are reported in Table \ref{table:ntu rgbd}, for the cross-subject, our ADR yields an obvious improvement over the softmax CE loss ($0.69\%$).
Additionally, the ADR achieves a noteworthy increase over the softmax CE loss in the protocol of cross-view ($0.55\%$). 
One can observe that the proposed ADR can yield consistent boosts in the Spatio-temporal architectures.

\begin{figure*}[!tb]
\begin{center}
\includegraphics[height=0.85\textheight]{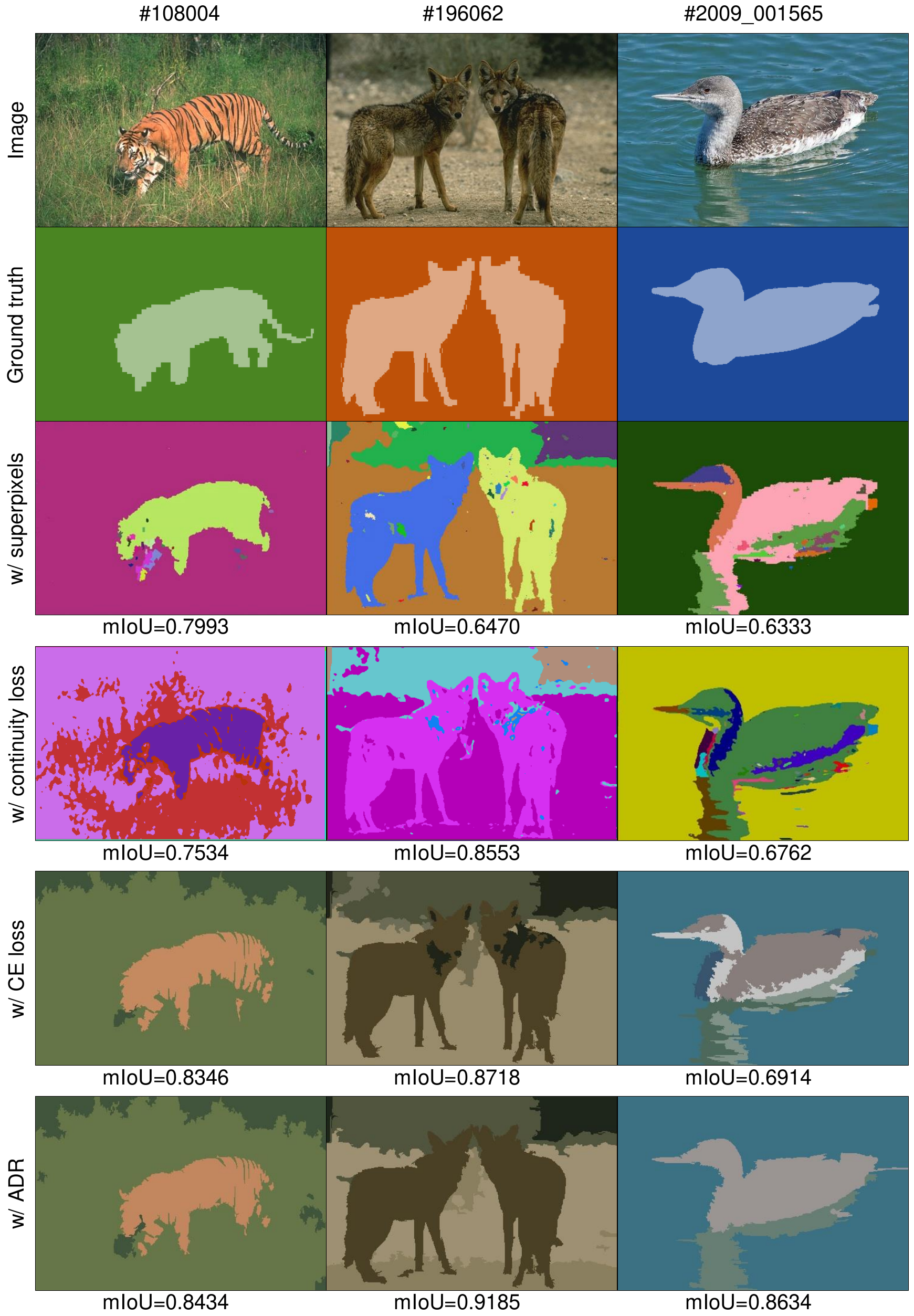}
\end{center}
    \caption{Qualitative results of the baseline and ours for unsupervised image segmentation on BSDS500 (\cite{arbelaez2010contour}) and PASCAL VOC 2012 (\cite{everingham2015pascal}).
    The original images are in the first row, the second consists of the ground truth images, the third contains the results of ``superpixels" (\cite{kanezaki2018unsupervised}), the fourth shows the results of continuity loss (\cite{kim2020unsupervised}), the fifth presents the results of CE, and our segmentation results are in the last row.
    We refer to \cite{kim2020unsupervised} to calculate the mIoU metrics and report it below the picture.
    Different segments are shown in different colors.}
\label{fig:segmentation}
\end{figure*}

\subsection{Unsupervised Image Segmentation}
As mentioned in Sec. \ref{sec:Introduction}, our ADR could be applied to several unsupervised learning tasks. 
Herein, we conduct an exploration namely unsupervised image segmentation by following a pioneer work (\cite{kim2020unsupervised}).
For this task, $\gamma$ is chosen from $\left [ 0.1, 0.5 \right]$ and $\tau$ is set as $\{ 7,9 \}$. 
Fig. \ref{fig:segmentation} presents the qualitative comparison between our ADR and the other methods. 
Both in the woof (second column) and waterfowl (third column) picture, more precise segments with various colors and textures are detected by our ADR.
More specifically, the model trained with ADR groups pixels within woof and waterfowl objects into a single category compared to the methods.

\section{Conclusion}
In this paper, we propose a practical regularization named adaptive discriminative regularization (ADR) for visual classification, verified on multiple datasets with a spectrum of the model architecture. It is built upon a hypothesis that we can leverage the data with similar semantics to calibrate the predicted likelihood in the target class and encourage it to be assertive, as data (e.g. images or videos) collected in the real-world are not ideally independent of each other.
Regarding this, ADR considers the intervention timing and gradient magnitude, exactly suitable for supervised visual classification tasks.
We give a general form and a simple solution to our method, and those qualities make it compatible with a wide range of visual applications.
Moreover, we demonstrate its availability and flexibility by conducting \textit{5} different visual classification tasks on over \textit{10} benchmarks. 
It also has potential value in objective function designing and other visual applications (e.g., collaborative learning (\cite{du2022i3cl}), multi-label learning (\cite{wu2018multi}) and clustering (\cite{castellano2022deep})).

\section*{Appendix} 
Our adaptive discriminative regularization loss for one sample can be written as 
\begin{equation}
    \label{BG_sample}
    \begin{aligned}
    {\cal L}_d({\tilde y}_i) & = \prod_{j=1}^{\tau} \frac{1}{\sqrt{2\pi\varphi_{i}}} \text{exp} \{-\frac{1}{2\varphi_{i}}{\hat y}^2_{ij}\}. \\
    & = \prod_{j=1}^{\tau} {\cal F}_j({\hat y}_{ij}), \\
    \end{aligned}
\end{equation}
where $\varphi_{i}$ is a function of ${\tilde y}_{i}$, ${\hat y}_{i}$ is generated by a non-linear function $\text{TopK}({\tilde y}_{i})$.
The function ${\cal F}_j({\hat y}_{ij})$ in Eq. \ref{BG_sample} can be denoted as
\begin{equation}
    \label{BG_sample-1}
    \begin{aligned}
    {\cal F}_j({\hat y}_{ij}) & = {\cal H}({\hat y}_{ij}){\cal E}({\hat y}_{ij}),
    \end{aligned}
\end{equation}
where ${\cal H}({\hat y}_{ij})$ is called the base measure function ``$ \frac{1}{\sqrt{2\pi\varphi_{i}}}$", ${\cal E}({\hat y}_{ij})$ is named the exponential term ``$ \text{exp}\{-\frac{1}{2\varphi_{i}}{\hat y}^2_{ij}\}$".

In the backward propagation, $ \frac{\partial {\cal L}_d({\tilde y}_i)}{\partial {\tilde y}_{i}} $ can be calculated  with
\begin{equation}
    \label{BG_sample-2}
    \begin{aligned}
    \frac{\partial {\cal L}_d({\tilde y}_i)}{\partial {\tilde y}_{i}} & = \sum_{j=1}^{\tau} \left [ {\cal F}'_j({\hat y}_{ij}) \prod_{m\neq j}^{\tau} {\cal F}_m({\hat y}_{ij}) \right ],\\
    \end{aligned}
\end{equation}
The derivative function ${\cal F}'_j({\hat y}_{ij}) $ in Eq. \ref{BG_sample-2} can be computed with 
\begin{equation}
    \label{BG_sample-3}
    \begin{aligned}
    {\cal F}'_{j}({\hat y}_{ij}) & = {\cal H}'({\hat y}_{ij}){\cal E}({\hat y}_{ij}) + {\cal E}'({\hat y}_{ij}){\cal H}({\hat y}_{ij}), \\
    \end{aligned}
\end{equation}
In Eq. \ref{BG_sample-3}, ${\cal H}'({\hat y}_{ij}) $ and ${\cal E}'({\hat y}_{ij}) $ can be calculated by
\begin{equation}
    \label{BG_sample-4}
    \begin{aligned}
    \frac{\partial {\cal H}({\hat y}_{ij})}{\partial {\hat y}_{ij}} & = \frac{1}{\sqrt{2\pi}}(-\frac{1}{2}\varphi_{i}^{-\frac{3}{2}})\varphi_{ij}' \\
    & = -\frac{\varphi_{ij}'}{2\varphi_{i}} {\cal H}({\hat y}_{ij}),\\
    \frac{\partial {\cal E}({\hat y}_{ij})}{\partial {\hat y}_{ij}} & = \left[ \frac{-{\hat y}_{ij}^2}{2\varphi_{i}} \right ]'{\cal E}({\hat y}_{ij}) \\
    & = \left [ \frac{{\hat y}_{ij}^2\varphi_{ij}'-2{\hat y}_{ij}\varphi_{i}}{2\varphi_{i}^2}\right ] {\cal E}({\hat y}_{ij}).\\
    \end{aligned}
\end{equation}
Putting Eq. \ref{BG_sample-4} into Eq. \ref{BG_sample-3}, ${\cal F}'_{j}({\hat y}_{ij}) $ can be rewritten as
\begin{equation}
    \label{BG_sample-5}
    \begin{aligned}
    {\cal F}'_{j}({\hat y}_{ij}) & = \left[  \frac{{\hat y}_{ij}^2\varphi_{ij}'-2{\hat y}_{ij}\varphi_{i}}{2\varphi_{i}^2} - \frac{\varphi_{ij}'}{2\varphi_{i}} \right ] {\cal F}_{j}({\hat y}_{ij}) \\ 
    & = \left[  \frac{{\hat y}_{ij}^2\varphi_{ij}'-2{\hat y}_{ij}\varphi_{i} - \varphi_{i}\varphi_{ij}'}{2\varphi_{i}^2} \right ] {\cal F}_{j}({\hat y}_{ij}),\\
    \end{aligned}
\end{equation}
Then, putting Eq. \ref{BG_sample-5} into Eq. \ref{BG_sample-2}, $ \frac{\partial {\cal L}_d({\tilde y}_i)}{\partial {\tilde y}_{i}} $ can be rewritten as 
\begin{equation}
    \label{BG_sample-6}
    \begin{aligned}
    & \frac{\partial {\cal L}_d({\tilde y}_i)}{\partial {\tilde y}_{i}} = \sum_{j=1}^{\tau} \left [ {\cal F}'_j({\hat y}_{ij}) \prod_{m\neq j}^{\tau} {\cal F}_m({\hat y}_{ij}) \right ]\\
    & = \sum_{j=1}^{\tau} \left [ \left ( \frac{{\hat y}_{ij}^2\varphi_{ij}'-2{\hat y}_{ij}\varphi_{i} - \varphi_{i}\varphi_{ij}'}{2\varphi_{i}^2} \right )\prod_{j=1}^{\tau}{\cal F}_m({\hat y}_{ij}) \right ]\\
    & = \sum_{j=1}^{\tau} \left [ \left ( \frac{{\hat y}_{ij}^2\varphi_{ij}'-2{\hat y}_{ij}\varphi_{i} - \varphi_{i}\varphi_{ij}'}{2\varphi_{i}^2} \right ) {\cal L}_d({\tilde y}_i)\right ],
    \end{aligned}
\end{equation}
where $\varphi_{i}' $ is the partial derivative function $\varphi_{i}$ with respect to ${\hat y}_{ij} $. We refer to \cite{sen2005n} and \cite{guariglia2021fractional}, $\varphi_{i}' $ can be computed with
\begin{equation}
    \label{BG_sample-7}
    \begin{aligned}
    \frac{\partial \varphi_{i}}{\partial  {\hat y}_{ij}} & = - \frac{\varphi_{i}+\text{log}({\hat y}_{ij})}{1-{\hat y}_{ij}}. \\
    \end{aligned}
\end{equation}

We also give the derivative function of entropy ${\cal L}'_e(p) $ for binary classification.
It can be calculated by 
\begin{equation}
\label{eq:a_d_hloss}
\begin{aligned}
\frac{\partial {\cal L}_e(p)}{\partial p} & = -\left [ log(p)-log(1-p)\right]\\
 & = log(\frac{1-p}{p}).
\end{aligned}
\end{equation}

\bibliography{sn-bibliography}

\section*{Data Availability} 
The datasets used during and analyzed during the current study are available in the following public domain resources: 
\begin{itemize}
	\item https://image-net.org/index.php;
	\item https://www.cs.toronto.edu/$\sim$kriz/cifar.html;
    \item https://www2.eecs.berkeley.edu/Research/\\Projects/CS/vision/grouping/resources.html;
    \item https://www.kaggle.com/c/challenges-in-\\representation-learning-facial-expression-\\recognition-challenge/data;
    \item http://www.cbsr.ia.ac.cn/english/CASIA-WebFace-Database.html;
    \item https://www.robots.ox.ac.uk/$\sim$vgg/data/\\flowers/102/;
    \item http://host.robots.ox.ac.uk/pascal/VOC/;
    \item http://vis-www.cs.umass.edu/lfw/;
    \item http://whdeng.cn/CALFW/index.html;
    \item http://whdeng.cn/CPLFW/index.html;
    \item https://ibug.doc.ic.ac.uk/resources/agedb/;
    \item http://www.cfpw.io/;
    \item http://rose1.ntu.edu.sg/datasets/\\actionrecognition.asp;
\end{itemize} 
The models and source data generated during and analyzed during the current study are available from the corresponding author upon reasonable request.

\section*{Acknowledgments} 
This work was supported by the National Natural Science Fund of China (62076184, 61673299, 61976160, 62076182), in part by Shanghai Innovation Action Project of Science and Technology (20511100700) and by  Shanghai  Natural Science Foundation (22ZR1466700); and in part by  Shanghai Municipal Science and Technology Major Project (2021SHZDZX0100) and the Fundamental Research Funds for the Central Universities.
Thanks to Xiaopeng Ji\footnote{Xiaopeng Ji is with the State Key Lab of CAD\&CG, Zhejiang University, China. (email: xp.ji@cad.zju.edu.cn)} and Xinyang Jiang\footnote{Xinyang Jiang is with the Microsoft Research Asia (Shanghai), Shanghai, China. (email: xinyangjiang@microsoft.com)} for their help with this work.

\end{document}